\documentclass{article} 
\usepackage{iclr2027_conference,times}

\usepackage{amsmath,amsfonts,bm}

\def\eqref#1{equation~\ref{#1}}

\def\1{\bm{1}}

\DeclareMathAlphabet{\mathsfit}{\encodingdefault}{\sfdefault}{m}{sl}
\SetMathAlphabet{\mathsfit}{bold}{\encodingdefault}{\sfdefault}{bx}{n}

\usepackage{hyperref}
\usepackage{url}
\usepackage{graphicx}
\usepackage{booktabs}

\title{EverMine: Dissecting the Self-Evolution of Research Capabilities in Long-Horizon Alpha Research}

\author{
Siyuan Li\textsuperscript{1,*}
\quad
Jiangfeng Zhang\textsuperscript{1,*}
\quad
Rui Yao\textsuperscript{1,*}
\quad
Weihua Qiu\textsuperscript{2}
\quad
Mingyang Xu\textsuperscript{3}
\quad
Zixuan Yuan\textsuperscript{1,\textdagger}
\\[6pt]
\textsuperscript{1}The Hong Kong University of Science and Technology (Guangzhou) \\
\textsuperscript{2}The Hong Kong Polytechnic University \\
\textsuperscript{3}Singapore Management University
\\[4pt]
\textsuperscript{*}Equal contribution.
\qquad
\textsuperscript{\textdagger}Corresponding author.
\\[4pt]
\texttt{\{sli974,jzhang67,ryao663\}@connect.hkust-gz.edu.cn} \\
\texttt{weihua.qiu@connect.polyu.hk}
\quad
\texttt{mingyang.xu.2026@phdacc.smu.edu.sg} \\
\texttt{zixuanyuan@hkust-gz.edu.cn}
}

\iclrpreprintcopy 
\begin{document}

\maketitle

\begin{abstract}
Self-evolving agents aim to turn research feedback into reusable skills, tools, and research rules. Whether these explicitly accumulated capabilities continue to improve later research requires controlled evaluation. Long-horizon alpha discovery provides a state-dependent setting for this question. Once a new factor enters the portfolio, it changes the predictive information already covered. The value of the same candidate factor or research experience may therefore change over time.

We introduce \textbf{EverMine}, an empirical framework for studying the self-evolution of research capabilities in long-horizon alpha discovery. EverMine decomposes the research state into research history ($ \mathrm{Hist} $), the current factor portfolio ($ \mathrm{Frontier} $), and reusable research capabilities ($ \mathrm{Cap} $). Under matched resource limits, we compare complete research runs with either fixed or continuously evolving $ \mathrm{Cap} $. We also replace $ \mathrm{Cap} $ while holding $ \mathrm{Hist} $ and $ \mathrm{Frontier} $ fixed to estimate the value of accumulated capabilities for subsequent research. Finally, we combine full trajectories with historical-state replay to study how experience-based rules change candidate selection and the resulting portfolio outcomes.

Across 18 long-horizon trajectories, end-to-end comparisons show no consistent gain from $ \mathrm{Cap} $ evolution. Across 48 continuation branches derived from shared $ \mathrm{Hist} $ and $ \mathrm{Frontier} $ states, accumulated $ \mathrm{Cap} $ also does not consistently outperform the initial $ \mathrm{Cap} $. Behavioral analysis shows that agents can still obtain portfolio improvements by tuning parameters of existing factor structures. Replays of two complete screening batches show that experience-based screening can retain high-value candidates while missing other candidates with positive marginal value. When all screened-out factors are then submitted sequentially, the final portfolio IC decreases slightly in both batches.

The replay results also show that the marginal value of a candidate changes as the portfolio is updated and as its submission position changes. Missed opportunities from experience-based screening therefore need to be evaluated in the actual research state. These results suggest that self-evolving research capabilities should be assessed from three complementary views: end-to-end research outcomes, the conditional value of accumulated capabilities, and the consequences of using experience in concrete research decisions.
\end{abstract}

\section{Introduction}

Long-horizon alpha research aims to keep finding new predictive value as the factor portfolio evolves. An alpha factor (Alpha) is constructed from market data such as historical prices and volumes and is used together with existing factors for prediction. The value of a new factor therefore depends both on its own predictive power and on how much new information it adds to the current portfolio\citep{yu2023generating}. As useful factors enter the portfolio, earlier discoveries change the remaining search space. The same candidate can have different marginal value under different portfolios, and a previously useful research direction can lose value as the portfolio changes. After controlling for existing factors, many new factors provide limited independent information, especially when the current portfolio already covers substantial predictive structure\citep{feng2020taming,green2017characteristics}. A long-horizon research agent must therefore keep identifying information that is still missing from the current portfolio and adapt its research accordingly.

Self-evolving agents aim to turn interaction and execution feedback into experience that can guide later behavior\citep{selfevolve2025,zhang2026agentic}. We study the explicit self-evolution of reusable research capabilities. The agent organizes research feedback into persistent, editable, and reusable skills, tools, and research rules that can guide later work. In long-horizon alpha research, new discoveries change the current portfolio, so the state in which an experience is formed can differ from the state in which it is later reused. This paper asks: can explicitly accumulated research capabilities continue to improve subsequent research on an evolving predictive frontier?

Prior work has enabled agents to autonomously propose, test, and revise alpha factors\citep{alphaagent2025,alphaagentevo2026}, and to accumulate experience, reuse skills, or update research workflows across multiple rounds\citep{factorminer2026,guo2026aqua,luo2026harness}. AlphaMemo and Agora further use component comparisons to study process memory and skill resources\citep{alphamemo2026,li2026trading}. During continual research, history, the portfolio, and capabilities evolve together. A comparison only at the final endpoint mixes the effects of all three. The additional value of accumulated capabilities therefore requires a comparison from the same research state. Experience-based rules also determine which candidates receive direct evaluation, leaving early-screened opportunities without feedback. Understanding the role of research capabilities in long-horizon research thus requires separating overall system performance from the conditional value of accumulated capabilities, while also tracing which research decisions are changed by experience.

\begin{figure}[t]
    \centering
    \includegraphics[width=\linewidth]{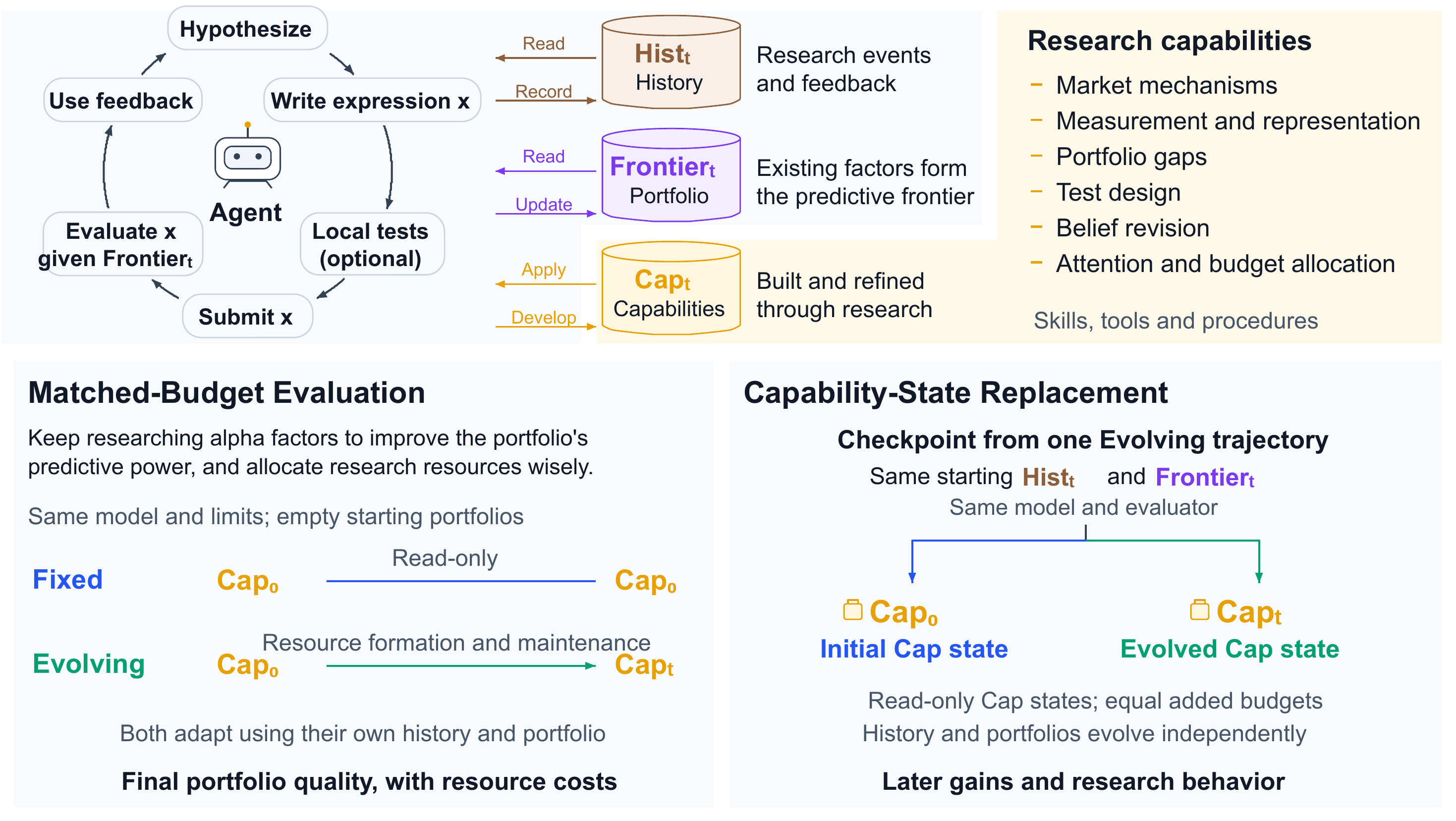}
    \caption{
    \textbf{EverMine research loop and evaluation framework.}
    The top panel shows three states in continual alpha research: research history (\(\mathrm{Hist}_t\)), the current factor portfolio (\(\mathrm{Frontier}_t\)), and reusable research capabilities (\(\mathrm{Cap}_t\)).
    The bottom panel shows two complementary evaluations: an end-to-end comparison of Fixed and Evolving under matched resource limits, and capability-state replacement from the same \(\mathrm{Hist}_t\) and \(\mathrm{Frontier}_t\).
    }
    \label{fig:evermine-overview}
\end{figure}

We introduce EverMine, a controlled empirical framework for studying self-evolving research capabilities in long-horizon alpha discovery. Figure~\ref{fig:evermine-overview} shows the design. EverMine explicitly separates research history (\(\mathrm{Hist}\)), the current factor portfolio (\(\mathrm{Frontier}\)), and reusable research capabilities (\(\mathrm{Cap}\)). Under matched resource limits, we compare a system with fixed initial $ \mathrm{Cap} $ (Fixed) against one with explicit $ \mathrm{Cap} $ evolution (Evolving). We then replace $ \mathrm{Cap} $ while holding \(\mathrm{Hist}\) and \(\mathrm{Frontier}\) fixed to test the effect of accumulated capabilities on subsequent research. We also combine full research trajectories with historical-state replay to examine how experience changes candidate-factor selection and the resulting portfolio outcomes.

Our experiments do not show consistent end-to-end gains from explicit $ \mathrm{Cap} $ evolution. Replacing $ \mathrm{Cap} $ at matched research states also does not show a consistent advantage for accumulated capabilities. We then examine the portfolio consequences of concrete research choices. Parameter tuning of existing factor structures contributes a positive net improvement in 16 of 18 trajectories, showing that familiar structures can remain useful targets for further research. Historical-state replay of two complete screening batches shows that experience-based screening can retain high-value candidates while missing other candidates with positive marginal value. After all screened-out factors are sequentially submitted, the final portfolio IC decreases slightly in both batches. These findings show that the usefulness of an experience-based rule depends on the current portfolio and should be evaluated through the choices it changes and the outcomes that follow.

Our main contributions are:

\begin{itemize}
    \item We introduce EverMine, which separately records research history, the current factor portfolio, and reusable research capabilities. By combining end-to-end comparisons with capability replacement at matched states, EverMine separates the overall system effect of capability evolution from the conditional value of accumulated capabilities.

    \item We combine full-trajectory statistics with historical-state replay to study continued search around existing factor structures, opportunities missed by experience-based screening, and portfolio outcomes after sequential submissions. These analyses connect concrete research choices to realized portfolio improvements and provide evidence on how reusable research capabilities are used in practice.
\end{itemize}

\section{Related Work}

\paragraph{Alpha Discovery on an Evolving Predictive Frontier.}
Automated alpha discovery has expanded from generating individual candidate factors to joint search around an existing factor set. AlphaGen includes complementarity among factors in its portfolio objective and uses joint predictive performance to guide expression generation\citep{yu2023generating}. AlphaQCM models search returns under a changing factor pool and sparse feedback, while AlphaSAGE uses structure-aware generation to explore diverse candidate factors\citep{zhu2025alphaqcm,chen2026alphasage}. LLM-based research agents further integrate financial hypotheses, candidate construction, and feedback-driven revision into multi-round research\citep{alphaagent2025,alphaagentevo2026,quantaalpha2026}. EverMine studies how research capabilities are reused while this predictive portfolio continues to evolve.

\paragraph{Reusable Experience and Self-Evolving Research Agents.}
Reflexion and ExpeL extract reusable language-based experience from execution feedback and past trajectories\citep{shinn2023reflexion,zhao2024expel}. ACE organizes context as an evolving strategy playbook that can be accumulated and revised over time\citep{zhang2026agentic}. In alpha and quantitative research, FactorMiner and AlphaMemo use skills, experience memory, and local search evidence to guide later exploration\citep{factorminer2026,alphamemo2026}. AQuA, AutoScientist-Quant, Agora, and HASE extend self-evolution to research organization, skill resources, evaluation rules, or execution frameworks\citep{guo2026aqua,li2026autoscientist,li2026trading,luo2026harness}. We focus on what these resources do after they have been formed and how experience-based judgments should be revised as the portfolio changes.

\paragraph{Evaluating Adaptation across Changing Research States.}
As evaluation moves from static candidate factors to continual research, prior work has begun to separate the final research product from the research process itself. AlphaMemo uses component comparisons to study the effect of process memory and search ledgers. Agora compares frozen skill libraries and conditions that are initialized with previously discovered scoring methods\citep{alphamemo2026,li2026trading}. Agentic Empirical Asset Pricing repeatedly reruns a full discovery process across historical dates to test whether an autonomous research system can continue producing new information\citep{pan2026agentic}. EverMine focuses on the evolving predictive frontier within a single long-horizon research trajectory. It combines end-to-end comparisons under matched resource limits with capability-state replacement from the same \(\mathrm{Hist}\) and \(\mathrm{Frontier}\), and uses historical-state replay to trace how experience-based rules change candidate selection and portfolio outcomes.

\section{Problem Setup}

\subsection{Portfolio-Conditioned Alpha Discovery}

EverMine provides an automated alpha-discovery environment for cross-sectional return prediction. The agent reads the current factor portfolio, past experiments, and platform feedback. It then proposes financial hypotheses, constructs expressions in a shared factor language, and submits candidate factors to the evaluator. The loop is illustrated in the upper-left part of Figure~\ref{fig:evermine-overview}.

In this paper, \emph{long-horizon} research means many rounds of experiments within the same research task, with shared history and a continuously updated portfolio. The data range and evaluation rules remain fixed during a run, while research history and the factor portfolio change with the agent's actions.

We measure research progress by portfolio predictive performance. A portfolio gain means an improvement in predictive quality and is measured by the change in portfolio IC. Let \(\mathrm{Frontier}_t\) denote the current factor portfolio and the predictive information it already covers, \(Q\) the portfolio evaluation function, and \(\operatorname{Update}\) the shared update algorithm. The marginal portfolio gain of a candidate factor \(x\) is
\[
\delta(x\mid \mathrm{Frontier}_t)=Q(\operatorname{Update}(\mathrm{Frontier}_t,x))-Q(\mathrm{Frontier}_t).
\]
In our experiments, \(Q\) is the time average of the per-bar cross-sectional Pearson correlation between the portfolio signal and the next 10-minute-bar return. This quantity is the portfolio Information Coefficient (IC). The factor portfolio has a fixed capacity. Following AlphaGen\citep{yu2023generating}, we set the maximum size to 30 and remove the member with the smallest absolute weight when the capacity is exceeded.

The value of the same candidate usually changes across portfolios, so \(\delta(x\mid \mathrm{Frontier}_t)\neq\delta(x\mid \mathrm{Frontier}_{t'})\) in general. The agent must therefore use its research experience together with the current portfolio when deciding which candidates are worth evaluating under limited resources.

\subsection{Research State with Capabilities}

At time \(t\), the state of a long-horizon research run is
\[
S_t=(\mathrm{Hist}_t,\mathrm{Frontier}_t,\mathrm{Cap}_t).
\]
\(\mathrm{Hist}_t\) contains the research history and its derived research map, including hypotheses, candidate factors, platform feedback, failure evidence, and conversational context. \(\mathrm{Frontier}_t\) is the current factor portfolio and the predictive information it already covers. \(\mathrm{Cap}_t\) is the explicit state of reusable research capabilities. It is implemented through skills, tools, and procedural documents. These resources contain research judgments and their conditions of use, resource indexes, and executable procedures that support later hypothesis formation, experiments, and feedback analysis. The agent can save, retrieve, and revise these resources.

For example, an agent may observe several momentum candidates that have predictive power alone but add little after entering the current portfolio. \(\mathrm{Hist}_t\) stores these observations, and \(\mathrm{Frontier}_t\) stores the portfolio at that time. The agent can then add a diagnostic rule to \(\mathrm{Cap}_t\): when similar results appear, first check how much the candidate overlaps with existing signals, then decide whether to tune the current formula structure or move to another direction. Later evidence can support this rule or lead the agent to narrow its scope and check additional properties such as candidate stability.

\section{Experimental Design}

\subsection{Fixed and Evolving Research Agents}

Fixed serves as the baseline condition. Its explicit capability state remains equal to the initial state \(\mathrm{Cap}_0\) throughout the trajectory. The agent still has access to the full continuous context, research history, current portfolio, shared research skills, and platform feedback. It can revise hypotheses, switch factor families, adjust its research pace, and use permitted local research tools.

Evolving uses the same configuration as Fixed and additionally allows the agent to update \(\mathrm{Cap}_t\) through a capability-management protocol. Both conditions can adapt their actions based on research history. Evolving can also save newly formed skills, tools, and procedures into $ \mathrm{Cap} $ and later retrieve, revise, or disable them.

\subsection{End-to-End Evaluation}

The end-to-end evaluation measures the overall difference between the Evolving and Fixed designs:
\[
\tau_{\mathrm{system}}
=
\mathbb{E}\!\left[
Q_{\mathrm{end}}(\mathrm{Evolving})
-
Q_{\mathrm{end}}(\mathrm{Fixed})
\right].
\]
Here, \(Q_{\mathrm{end}}\) is the realized portfolio quality at the end of research. \(\tau_{\mathrm{system}}\) measures the end-to-end change under matched resource limits, including the costs of creating and using capabilities. We record each realized resource consumption separately.

The main end-to-end comparison uses final portfolio quality in both the development and out-of-sample periods. We also record stepwise research trajectories over relative submission progress, improvements after the factor pool first reaches capacity, and resource use.

\subsection{Checkpoint-Based Capability-State Replacement}

Capability-State Replacement estimates the conditional value of an accumulated $ \mathrm{Cap} $ state for subsequent research. From a checkpoint on the same Evolving parent trajectory, we create two continuation conditions:
\[
(\mathrm{Hist}_t,\mathrm{Frontier}_t,\mathrm{Cap}_t),\qquad
(\mathrm{Hist}_t,\mathrm{Frontier}_t,\mathrm{Cap}_0).
\]
\(\mathrm{Cap}_t\) is the capability-state snapshot from the Evolving parent trajectory at that checkpoint. \(\mathrm{Cap}_0\) is the initial state supplied at the start of the Evolving condition. After branching, both $ \mathrm{Cap} $ states are frozen as read-only. Each branch then develops its own research history and portfolio through its subsequent actions.

The main effect is
\[
\tau_{\mathrm{cap}}
=
\mathbb{E}\!\left[\Delta Q\mid\mathrm{Cap}_t\right]
-
\mathbb{E}\!\left[\Delta Q\mid\mathrm{Cap}_0\right],
\]
where \(\Delta Q\) is the improvement in portfolio quality from the shared starting portfolio to the end of the continuation. \(\tau_{\mathrm{cap}}\) measures the subsequent value of the accumulated capability snapshot relative to the initial $ \mathrm{Cap} $.

The main analysis uses \(n\) Evolving parent trajectories from the same model. We create \(\mathrm{Cap}_t\) and \(\mathrm{Cap}_0\) branches at the 25\% and 75\% checkpoints of joint-budget progress. Each condition is repeated twice, producing \(8n\) continuation branches in total. A parent trajectory, checkpoint, and repetition index define a paired block containing both $ \mathrm{Cap} $ conditions.

Within each checkpoint, we first average the two continuations for each capability condition within each parent trajectory. We then compute the difference in subsequent gains between the two $ \mathrm{Cap} $ states and average equally across parent trajectories. Model settings are described in Section~\ref{sec:experimental-setup}, and checkpoint and budget definitions are given in Appendix~\ref{app:experiment-setup}.

\subsection{Research Behavior Analysis}

We first test whether continued research around existing formulas can improve the portfolio. For all long trajectories in both conditions, we classify factor submissions after the pool first reaches capacity into three categories: new-structure submissions, parameter-tuning submissions on existing structures, and repeated submissions of previously recorded factors. A new-structure submission uses a formula structure that has not appeared earlier in the same trajectory. A parameter-tuning submission keeps the data fields and operator structure unchanged but changes numerical parameters such as window length or exponent. For example, changing the window of a moving-average formula from 12 to 24 creates a new formula under an existing structure. A repeated submission uses exactly the same formula as an earlier submission, including the numerical parameters. Because \(\mathrm{Frontier}\) continues to change, resubmitting an old factor can test whether its value has changed under the current portfolio. Classification uses only records available before each submission. For each category, we sum realized changes in portfolio IC and retain negative gains.

We then test whether factors rejected by experience-based screening can still have value. We select batches with complete candidate sets, recorded screening reasons, and recoverable platform states. These include two batches from one Qwen Evolving trajectory and one batch from a Qwen Fixed trajectory. For each candidate, we restore the exact state at the screening decision and evaluate that candidate alone. We retain results for both accepted and screened-out candidates. For the two Evolving batches, we also restore the original state and sequentially submit all screened-out factors in their original order. The state is updated after every submission. We compare the final portfolio with the portfolio obtained by keeping the original decisions. The first replay measures opportunities missed at the original state, while the sequential replay measures the realized outcome of actually submitting the additional candidates. Results appear in Section~\ref{sec:behavior-results}; recovery procedures and additional diagnostics are in Appendix~\ref{app:decision-evidence}.

\subsection{Experimental Setup}
\label{sec:experimental-setup}

We evaluate a locally deployed Qwen3.8-27B~\citep{team2026qwen3} and DeepSeek-V4.1-Flash through the official DeepSeek API~\citep{deepseekai2026deepseek}. Their deployment identifiers are \texttt{qwen3.8-27b-272k-local} and \texttt{deepseek-flash}, respectively. Both models use a context window of 262,144 tokens. Qwen3.8-27B uses \texttt{xhigh} reasoning effort, while DeepSeek-V4.1-Flash uses \texttt{high}. The 18 long trajectories were run from September 20 to 24, 2026. Per-session contracts record the deployment identifier, runtime release, context limit, and reasoning setting. The agent runtime is DeepSeek Harness~\citep{deepseekai2026deepseeka}. Every independent research run, called a Session in the system logs, uses a separate container and file system. The agent conducts continual research in that environment and receives evaluation feedback on candidate factors and the current portfolio through a shared submission interface. The research protocol shared across conditions and the condition-specific capability interfaces are detailed in Appendix~\ref{app:research-protocol}.

The experiments use a point-in-time U60 universe from Binance Spot USDT markets with 10-minute market data. At the start of each month, the universe selects 60 trading pairs by turnover over the preceding 30 days. The development period is 2025-01-01 to 2025-12-31, and the out-of-sample (OOS) period is 2026-01-01 to 2026-06-30. The prediction target is the close-to-close return of the next 10-minute bar. The agent constructs factors using only the development period. The evaluator maintains a factor pool with capacity 30 under a fixed update rule and measures predictive quality by portfolio IC. For every OOS endpoint, we freeze the complete development-period portfolio state, including its members, directions, and weights, and use OOS labels only to compute its reported IC. The OOS results remain hidden from the research agent.

Each complete long-horizon trajectory uses a multi-resource budget with limits of USD 10 equivalent model cost, 1,000 platform submissions, 16 sandbox CPU core-hours, 10M output tokens, 10,000 successful research responses, 100 GiB cumulative writes, and 48 active wall-clock hours. Each capability-replacement branch receives additional limits of USD 3 model cost, 300 submissions, 4.8 sandbox CPU core-hours, 3M output tokens, 3,000 successful research responses, 30 GiB cumulative writes, and 14.4 active wall-clock hours. Resources are metered separately. The compared conditions use identical limits, and new research actions stop when any limit is reached. Resource definitions, safety recovery limits, and hardware are detailed in Appendix~\ref{app:experiment-setup}.

The end-to-end comparison contains 18 independent long trajectories: six Fixed and six Evolving runs for Qwen, and three Fixed and three Evolving runs for DeepSeek. Capability replacement uses the six Qwen Evolving trajectories as parent trajectories and creates 48 continuation branches. Full-trajectory behavioral statistics use all 18 long trajectories.

\section{Results and Analysis}

\subsection{End-to-End Performance in Long-Horizon Research}

\begin{figure*}[t]
    \centering
    \includegraphics[width=\textwidth]{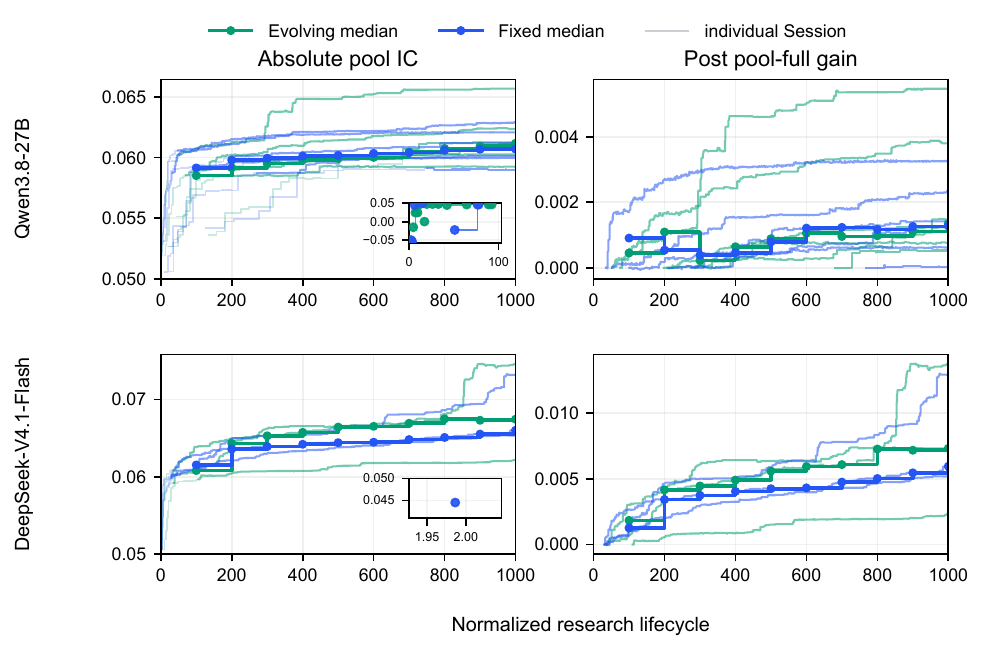}
    \caption{\textbf{Long-horizon research trajectories over relative submission progress.} Submission progress is normalized to 0--1000 within each trajectory. Results on fixed resource axes are reported in Appendix~\ref{app:analysis-details}.}
    \label{fig:long-horizon-performance}
\end{figure*}

The 18 long-horizon trajectories contain 530.5 active runtime hours and 8,774 candidate-factor evaluations, of which 8,770 have recorded portfolio outcomes. Figure~\ref{fig:long-horizon-performance} shows research progress over normalized submission position. Both models build an initial portfolio quickly, then continue replacing members, with some trajectories finding additional improvements later in the run. After the factor pool first reaches capacity, the mean net IC gains are \(0.001471 / 0.002202\) for Qwen Fixed / Evolving and \(0.007995 / 0.007782\) for DeepSeek. In total, 8,215 submissions, about \(93.6\%\), occur after the pool first reaches capacity. Most research therefore takes place while an existing portfolio is already present.

\begin{table*}[t]
\caption{\textbf{Complete end-to-end endpoint results.} IC values and intervals are multiplied by \(1000\). \(n_F\) and \(n_E\) are the numbers of independent trajectories. Differences are Evolving minus Fixed, with 95\% bootstrap intervals.}
\label{tab:primary-effects}
\centering
\small
\resizebox{\textwidth}{!}{
\begin{tabular}{lcccccc}
\toprule
Model & $n_F$ & $n_E$ & Mean Fixed IC & Mean Evolving IC & Mean IC diff. $\Delta$ & 95\% interval \\
\midrule
\multicolumn{7}{l}{\textbf{Development}} \\
Qwen3.8-27B & 6 & 6 & 60.902 & 61.643 & +0.741 & [-1.181, +2.681] \\
DeepSeek-V4.1-Flash & 3 & 3 & 68.166 & 68.071 & -0.095 & [-6.819, +6.624] \\
\addlinespace
\multicolumn{7}{l}{\textbf{OOS}} \\
Qwen3.8-27B & 6 & 6 & 88.324 & 87.932 & -0.391 & [-4.158, +3.180] \\
DeepSeek-V4.1-Flash & 3 & 3 & 100.779 & 99.141 & -1.637 & [-18.227, +13.089] \\
\bottomrule
\end{tabular}
}
\end{table*}

Table~\ref{tab:primary-effects} reports the mean final portfolio IC and between-condition differences. For Qwen, the development-period means are \(0.060902 / 0.061643\) for Fixed / Evolving. For DeepSeek, they are \(0.068166\) and \(0.068071\). In the OOS period, the Evolving mean is slightly lower than the Fixed mean for both models. All four difference intervals include zero. The current comparison therefore does not show a consistent gain from explicit $ \mathrm{Cap} $ evolution. Both conditions continue improving after the pool reaches capacity. We next isolate the additional value of accumulated $ \mathrm{Cap} $ from a shared research state. Resource analyses are reported in Appendix~\ref{app:analysis-details}.

\subsection{Conditional Value of Accumulated Capabilities}

Capability replacement restricts the comparison to the same starting research history and portfolio. The 48 continuation branches form 24 complete paired blocks. Within each checkpoint, we first average repetitions within each parent trajectory and $ \mathrm{Cap} $ condition, then average equally across parent trajectories. Table~\ref{tab:capability-set-replacement-effects} reports both the mean subsequent gain under each $ \mathrm{Cap} $ state and their difference.

\begin{table*}[t]
\caption{\textbf{Complete results for capability-state replacement.} IC gains and intervals are multiplied by \(1000\). Starting groups are paired blocks, and parent trajectories are the independent statistical units. The effect is \(\Delta Q(\mathrm{Cap}_t)-\Delta Q(\mathrm{Cap}_0)\), with 95\% bootstrap intervals.}
\label{tab:capability-set-replacement-effects}
\centering\small
\resizebox{\textwidth}{!}{
\begin{tabular}{llcccccc}
\toprule
Model & Checkpoint & Parent runs & Paired blocks & $\Delta Q(\mathrm{Cap}_0)$ & $\Delta Q(\mathrm{Cap}_t)$ & $\widehat{\tau}_{\mathrm{cap}}$ & 95\% interval \\
\midrule
\multicolumn{8}{l}{\textbf{Development}} \\
Qwen3.8-27B & 25\% & 6 & 12 & +2.034 & +1.628 & -0.405 & [-1.097, +0.119] \\
Qwen3.8-27B & 75\% & 6 & 12 & +0.500 & +0.542 & +0.042 & [-0.494, +0.573] \\
\addlinespace
\multicolumn{8}{l}{\textbf{OOS}} \\
Qwen3.8-27B & 25\% & 6 & 12 & +3.005 & +1.749 & -1.256 & [-3.107, +0.197] \\
Qwen3.8-27B & 75\% & 6 & 12 & +0.718 & +0.626 & -0.092 & [-2.048, +1.574] \\
\bottomrule
\end{tabular}
}
\end{table*}

At the 25\% checkpoint, the development-period gains are \(0.002034 / 0.001628\) for the initial / accumulated $ \mathrm{Cap} $. At the 75\% checkpoint, they are \(0.000500 / 0.000542\). Both $ \mathrm{Cap} $ states continue to improve the portfolio at both checkpoints. The mean difference between accumulated and initial $ \mathrm{Cap} $ is \(-0.000405\) at 25\% and \(+0.000042\) at 75\%. In OOS evaluation, both checkpoint differences favor the initial $ \mathrm{Cap} $. The intervals for both development and OOS differences include zero. Later branches also produce outputs that are closer to their own starting portfolios, and this persistence appears under both $ \mathrm{Cap} $ conditions. Additional analyses of rule use and late-stage persistence are in Appendix~\ref{app:cap-behavior}.

\subsection{Research Behavior Analysis}
\label{sec:behavior-results}

We next connect research choices to portfolio outcomes. We study continued attempts around existing formula structures and the effect of experience-based screening on factor submissions.

\paragraph{Parameter tuning on existing formula structures can still improve the portfolio.}
There are 8,215 submissions after the pool first reaches capacity, with usable gains for 8,211. Table~\ref{tab:research-decisions}A shows that new-structure submissions contribute the most in all four groups. Parameter-tuning submissions on existing structures account for 12.8\%--40.0\% of each group's net improvement. Their net contribution is positive in 16 of 18 trajectories, negative in one trajectory, and absent in one trajectory. For every group, the mean net contribution from parameter tuning remains positive after removing any single trajectory. Familiar formula structures can therefore retain useful portfolio value for further research.

\begin{table*}[t]
\caption{\textbf{Research choices and realized portfolio gains.} IC values are multiplied by \(1000\). A: submissions after the pool first reaches capacity are classified as new structures, parameter tuning of existing structures, or repeated submissions of previously recorded factors. Net gains are summed within each trajectory and then averaged equally within each group. Four repeated-submission outcomes are missing and are reported using known contributions. B: two complete screening batches from the same Qwen Evolving trajectory. ``Positive on original portfolio'' counts screened-out factors with \(\Delta\mathrm{IC}>10^{-6}\) when evaluated independently at the original state. Additional factors are then submitted in a fixed order, and the final portfolio is reported.}
\label{tab:research-decisions}
\centering
\small
\resizebox{\textwidth}{!}{
\begin{tabular}{llrrrr}
\toprule
\multicolumn{6}{l}{\textbf{A. Known net contributions after the pool first reaches capacity}} \\
Model & $ \mathrm{Cap} $ & New structures & \shortstack[r]{Parameter tuning\\of existing structures} & \shortstack[r]{Exact resubmissions of\\previously recorded factors} & \shortstack[r]{Runs with positive\\tuning contribution} \\
\midrule
Qwen & Fixed & $+0.932$ & $+0.588$ & $-0.049$ & 4/6 \\
Qwen & Evolving & $+1.657$ & $+0.567$ & $-0.022$ & 6/6 \\
DeepSeek & Fixed & $+5.910$ & $+2.009$ & $+0.076$ & 3/3 \\
DeepSeek & Evolving & $+6.736$ & $+0.996$ & $+0.050$ & 3/3 \\
\addlinespace
\multicolumn{6}{l}{\textbf{B. Screened-out factors: original-state gains and sequential-submission outcomes}} \\
Batch & Screened out & Positive on original portfolio & Original endpoint & Endpoint after additions & Endpoint diff. \\
\midrule
A & 3 & 2 & $61.14162$ & $61.12412$ & $-0.01749$ \\
B & 8 & 2 & $61.18655$ & $61.17988$ & $-0.00666$ \\
\bottomrule
\end{tabular}
}
\end{table*}

\paragraph{Experience-based screening can identify strong priorities while missing other valuable factors.}
We audit two complete screening batches from one Qwen Evolving trajectory. The batches contain 12 candidate factors, 11 of which were screened out because their local scores did not meet self-imposed thresholds. Batch A reused a threshold from the previous experiment, which was later written into $ \mathrm{Cap} $. Batch B explicitly referenced an existing $ \mathrm{Cap} $ procedure and combined it with a screening rule in the current experiment plan. We restore the original state and evaluate every candidate independently. Among the screened-out factors, four have positive marginal gains, six have negative gains, and one has a zero gain. All four positive gains exceed \(10^{-5}\). The accepted candidate in Batch A has a gain of \(0.00044635\), larger than every screened-out candidate in that batch, while the best screened-out candidate still has a gain of \(0.00014081\). In a separate audit of eight factors from a Qwen Fixed trajectory, both accepted and screened-out factors have negative gains. These cases separately test the priority selected by the rule and the opportunities it leaves unexplored.

We then restore the original states of the two Evolving batches and sequentially submit all screened-out factors in their original order, keeping each portfolio update. In Batch A, both comparison paths first submit the originally accepted factor, after which the additional-submission path submits three screened-out factors. In Batch B, we compare the original portfolio against the portfolio after sequentially submitting all eight screened-out factors. Of the four screened-out factors that have positive gains when evaluated independently on the original portfolio, three become negative when submitted in sequence because earlier submissions have already changed the portfolio. The additional set also contains factors whose original gains were negative or zero. As shown in Table~\ref{tab:research-decisions}B, the final IC after submitting every screened-out factor is lower than the original-decision endpoint by \(0.00001749\) in Batch A and \(0.00000666\) in Batch B. Independent value at the original state and realized value under sequential submission therefore answer different questions. Improving a screening rule requires considering both which factors deserve reevaluation and the portfolio outcome produced by those additional attempts. Rule versions, replay details, and additional analyses of repeated historical factors appear in Appendix~\ref{app:decision-evidence}.

\section{Limitations}

Capability-state replacement estimates the subsequent value of accumulated \(\mathrm{Cap}\) relative to the initial \(\mathrm{Cap}\). Both branches can still use past evidence through \(\mathrm{Hist}\). The separate effects of skills, tools, and procedural documents require additional interventions. The independent units in the main experiments are long trajectories or parent trajectories, and the current number of repetitions limits the precision of effect estimates. DeepSeek contributes three trajectories per condition and serves as a cross-model replication; its bootstrap intervals summarize the observed uncertainty but provide limited support for nominal 95\% coverage at this sample size.

OOS results cover all main trajectories and paired groups. Here, OOS means that the 2026 data and labels are held out from the agent's explicit research and evaluator loop. Because the models' pretraining corpora and temporal cutoffs are not fully observable, this split does not establish that the period was unseen during model training and should not be interpreted as post-training temporal generalization. This uncertainty is shared by Fixed and Evolving within each model but limits the scope of the OOS claim. The behavioral analyses are exploratory analyses conducted after inspecting the main results. The replay study covers verified complete batches, and the repeated-submission analysis uses factors that the agent actually chose to resubmit. Estimating screening rates across trajectories and long-run policy value will require broader controlled evaluation.

Our experiments use structured \(\mathrm{Cap}\), a fixed-capacity factor pool, and 10-minute cryptocurrency data. Other capability representations, portfolio update rules, asset classes, and frequencies provide natural settings for further tests.

\section{Conclusion}

EverMine studies whether explicitly accumulated reusable research capabilities continue to improve later research. End-to-end comparisons measure the full effect of explicit $ \mathrm{Cap} $ evolution. Capability replacement from a shared starting state measures the subsequent value of accumulated $ \mathrm{Cap} $. Behavioral analysis then examines the portfolio consequences of concrete research choices. Across 18 long trajectories and 48 continuation branches, we do not observe consistent improvement. Effect intervals in both development and OOS evaluation include zero.

Full-trajectory analysis shows that both new-structure submissions and parameter tuning of existing structures contribute to research gains. Parameter tuning has a positive net contribution in 16 of 18 trajectories. Replays of complete screening batches show that screening can retain the strongest candidate while still missing other candidates with positive marginal value. Yet sequentially submitting every screened-out candidate slightly lowers the final endpoint in both audited batches from the same trajectory. These results support evaluating three quantities separately: the priority selected by experience, the alternatives that experience rejects, and the realized portfolio outcome after continued research.

The Evolving protocol already requires the agent to record the basis, applicable context, and failure conditions of research judgments. Even under this design, neither the end-to-end comparison nor capability replacement shows consistent improvement. Future capability design and evaluation should connect recorded experience more directly to actual use and portfolio feedback. In particular, evaluation should track when an agent applies a judgment, which factors it continues or abandons because of that judgment, and how new evidence leads the agent to revise it.

\section*{Ethics and Reproducibility Statement}

\paragraph{Ethics Statement.}
This work is intended solely for academic research and discussion. It does not represent the views of any organization, institution, or individual, and it does not constitute investment advice.

\paragraph{Reproducibility Statement.}
The main text and appendices provide the experimental design and implementation details needed to reproduce the study. The market data used in this work are public and can be obtained from their official source. We will provide the necessary code during review and release the code publicly upon acceptance.

\subsection*{AI Use Statement}

Generative AI tools were used to assist with paper organization and language editing. The authors manually reviewed all AI-assisted content and remain fully responsible for the manuscript.

\bibliography{references}
\bibliographystyle{iclr2027_conference}

\appendix
\section{Extended Related Work}
\label{app:extended-related-work}

\subsection{Portfolio-Conditioned Alpha Discovery and Research Processes}

Financial research has long studied how much incremental information a new signal provides beyond existing predictors. In \emph{Taming the Factor Zoo}, Feng, Giglio, and Xiu test new pricing factors after controlling for a large set of existing factors. They find that many candidate factors can be absorbed by known factors and that only a subset provides independent information\citep{feng2020taming}. Green, Hand, and Zhang jointly evaluate a large set of firm characteristics and similarly find that relatively few characteristics provide stable, independent return-predictive information\citep{green2017characteristics}. These studies motivate evaluating a new factor by its incremental value relative to information that is already available.

Formula-based alpha discovery makes this complementarity part of the search objective. AlphaGen trains an expression-generation policy using the joint predictive performance of a factor set, so the search directly targets value added to existing factors\citep{yu2023generating}. AlphaQCM models alpha discovery as a sequential decision problem with a changing factor pool and sparse feedback, and models the distribution of search returns\citep{zhu2025alphaqcm}. AlphaSAGE uses a structure-aware generative flow network to broaden structural exploration and improve candidate diversity\citep{chen2026alphasage}. These methods address factor complementarity, search-space coverage, and continued search under a changing factor pool from different directions.

LLM-based agents extend alpha search into a research process that includes hypothesis formation, expression construction, experimental feedback, and revision. AlphaAgent translates financial hypotheses into testable expressions and organizes exploration with constraints such as novelty and complexity\citep{alphaagent2025}. QuantaAlpha treats the research trajectory of hypotheses, expressions, implementation, and feedback as the object of evolution\citep{quantaalpha2026}. R\&D-Agent-Quant jointly organizes factor and prediction-model research, while AlphaAgentEvo uses agent reinforcement learning to improve multi-round tool use, planning, and reflection\citep{rndagent2025,alphaagentevo2026}. These systems increasingly move the research target from one-shot candidate generation to a continuing research process.

\subsection{Reusable Experience and Self-Evolving Research Agents}

A central mechanism in adaptive and self-evolving agents is to let past interactions continue to affect later behavior. Reflexion converts execution feedback into language-based reflections that can guide future attempts\citep{shinn2023reflexion}. ExpeL extracts reusable experience rules from multiple successful and failed trajectories\citep{zhao2024expel}. Agentic Context Engineering (ACE) further organizes context into a continually updated strategy playbook, maintained through incremental generation, reflection, and curation\citep{zhang2026agentic}. These methods provide different ways to convert concrete interaction histories into reusable resources.

In alpha research, such experience directly affects research directions and candidate selection. FactorMiner combines modular skills with an updateable experience memory that preserves useful search patterns and failures in highly redundant directions\citep{factorminer2026}. AlphaMemo records local modifications around parent factors, evaluation results, and search genealogy, and uses confidence-aware process memory and failure evidence to guide later search\citep{alphamemo2026}. In these systems, experience affects where the next search starts, which modifications are worth trying, and which directions should receive lower priority.

More open-ended autonomous research systems can also change how research itself is organized. AQuA preserves evidence across rounds of factor and model research and recursively chooses later research based on prior results\citep{guo2026aqua}. AutoScientist-Quant uses current progress and remaining budget to decide whether to improve an existing result, combine earlier results, or switch to a new research path\citep{li2026autoscientist}. Agora continually updates executable skills and internally developed scoring methods\citep{li2026trading}, while Harness-Aware Self-Evolving (HASE) expands the object of evolution to model weights, the agent harness, and task solutions\citep{luo2026harness}. At the same time, \emph{Your Agent May Misevolve} shows that self-evolution can also create and reinforce undesirable behavior patterns\citep{shao2026your}.

\subsection{Evaluation of Adaptation under Evolving Research States}

Evaluating continual adaptive research requires studying both the outcome of the full process and the effect of accumulated experience or capabilities on later actions.

Prior alpha-research systems have begun to use component comparisons to isolate experience resources. AlphaMemo removes additional process memory while keeping factor quality, search genealogy, and historical records available, allowing it to estimate the contribution of process memory to search behavior and outcomes\citep{alphamemo2026}. Agora compares continually evolving and frozen skill libraries. It also initializes some control conditions with scoring methods discovered during earlier research to study the effect of different research products\citep{li2026trading}. These experiments provide controlled evidence on memory and skill components and show that different information sources in long-horizon research require different interventions.

Another line of work evaluates the complete discovery process across time. \emph{Agentic Empirical Asset Pricing} treats a repeatedly executable autonomous discovery system as the evaluation target. By rerunning the research process at consecutive historical dates, it tests whether the system can continue producing discoveries with OOS predictive power and incremental information, and studies components such as research logs and experience summaries\citep{pan2026agentic}. This view emphasizes that an autonomous research system should be evaluated both by the product of one run and by whether the process continues to generate useful discoveries under later data and research states.

General long-horizon agent research also provides controlled methods for separating historical conditions from later behavior. Laban et al. vary whether the same information is presented as a single-turn input or as a multi-turn interaction, and study how previous interaction affects later reliability in long conversations\citep{laban2026llms}. Sinha et al. vary the fraction of earlier execution errors in controlled tasks to separate the effect of long context from self-conditioning on the agent's own past actions\citep{sinha2026illusion}. MemoryAgentBench decomposes long-term memory into retrieval, test-time learning, long-horizon understanding, and selective forgetting, and measures these abilities through incremental multi-turn interactions\citep{hu2026evaluating}. These studies offer methodological references for analyzing adaptation by controlling the research state or interaction history.

\subsection{Evaluating Alpha Research and Financial Evidence}

Alpha research can be evaluated at the level of candidate factors, factor sets, or final investment outcomes. AlphaBench decomposes formula-based alpha research into candidate generation, quality judgment, and feedback-driven search, providing component-level evaluation of LLM capabilities in local research tasks\citep{luo2026alphabench}. BacktestBench evaluates automated quantitative backtesting through tasks such as metric calculation, asset selection, strategy selection, and parameter confirmation\citep{wang2026backtestbench}. These benchmarks mainly study local research and backtesting capabilities. Continual alpha discovery also requires evaluating the signal set produced over many research rounds.

Existing automated alpha systems use different outcome measures. FactorMiner evaluates candidate predictive quality and redundancy, counts discoveries that pass its screening criteria, and compares the portfolio performance of discovered factors\citep{factorminer2026}. AlphaGen directly organizes search around the joint predictive performance of a factor set\citep{yu2023generating}. AQuA additionally reports investment-simulation outcomes beyond research artifacts\citep{guo2026aqua}. These metrics answer different questions. Single-factor predictive quality measures whether a candidate contains return information. The number of accepted discoveries measures research output under a particular quality criterion. Portfolio predictive quality measures how much predictive information multiple signals provide when used together. Investment backtests additionally depend on portfolio construction, transaction costs, and other implementation choices.

Repeated testing, result selection, and temporal leakage can also affect the interpretation of financial evidence. Bailey and L\'opez de Prado's Deflated Sharpe Ratio analyzes the effects of multiple testing, selection bias, backtest overfitting, and non-normal returns on performance statistics\citep{bailey2014deflated}. McLean and Pontiff document that published return predictors often lose predictive power out of sample and after publication\citep{mclean2016does}. For LLM-based financial research systems, DeepFund further notes that historical financial data and related research may have appeared in model pretraining corpora, so historical time-split evaluation should also consider the boundary of pretraining information\citep{li2025time}.

Given these distinctions, EverMine reports the continual search process on the development period separately from OOS evaluation of the frozen final portfolio. Development-period results show how the agent obtains marginal gains on an evolving \(\mathrm{Frontier}_t\) and how research behavior changes. OOS results test the predictive performance of the final research product on data that were not used by the research agent during search.

\section{Experimental Setup}
\label{app:experiment-setup}

\subsection{Research Conditions and Protocol}
\label{app:research-protocol}

\subsubsection{Shared Research Protocol}

Fixed, Evolving, and capability-replacement runs share the same research task, data, common research materials, factor-submission interface, platform evaluation feedback, and resource accounting. They also use the same pre-specified research-status mechanism. When an agent calls \texttt{evermine-submit runtime-status}, the platform constructs a read-only snapshot from the current objective, portfolio version, resource counters, and submission queue. If these measurements are available, the response includes the same protocol reminder: submitting a valid hypothesis is itself a research test, local metrics or a complete mechanism proof are not prerequisites for submission, and a self-defined local threshold cannot replace the platform evaluation loop. The reminder code and text were frozen in the common runtime release before the reported experiments.

The reminder is generated by the status query alone. Its trigger does not inspect the experimental condition, candidate content, $ \mathrm{Cap} $ state, or whether the agent has adopted a local screening rule. The mechanism is equally available in all conditions, while the timing of status queries and the agent's subsequent use of the returned fields remain part of its research behavior. Consequently, trajectories can receive or act on the reminder at different times without any trajectory-specific message being added by the researchers.

\subsubsection{Fixed Condition}

Within this shared protocol, the Fixed condition loads the initial capability state \(\mathrm{Cap}_0\), whose contents and version remain read-only throughout the run. The agent can still read the full continuous context, update \(\mathrm{Hist}_t\), respond to changes in \(\mathrm{Frontier}_t\), and use permitted local research tools. Hypotheses, candidate factors, and feedback produced during the run enter the research history, while the explicit \(\mathrm{Cap}\) stays at its initial version. This design preserves research adaptation based on historical information while providing a fixed reference for explicit $ \mathrm{Cap} $ evolution in the Evolving condition.

\subsubsection{Evolving Condition}

Evolving keeps the shared protocol and additionally receives a capability taxonomy, a writable capability state \(\mathrm{Cap}_t\), and interfaces for versioning and retrieval. The agent can create, save, load, revise, and disable reusable research resources. The management protocol requires each resource to record its applicable context, the research choice being considered, the action it is expected to change, supporting and opposing evidence, failure conditions, and where it should be used next. Specific factors, one-off thresholds, and local experimental facts are kept mainly in \(\mathrm{Hist}_t\). \(\mathrm{Cap}_t\) stores research judgments and execution methods intended for reuse across later experiments or research stages.

The capability taxonomy helps the Evolving agent organize accumulated research experience and resources. The structured Evolving condition provides six dimensions for research judgment and one engineering-support dimension. A resource can span multiple dimensions, and the agent can extend the taxonomy with additional research judgments.

\begin{table*}[t]
\caption{Capability dimensions in the Evolving condition and their intended changes to research behavior.}
\label{tab:self-evolution-dimensions}
\begin{center}
\small
\begin{tabular}{p{0.14\linewidth}p{0.38\linewidth}p{0.38\linewidth}}
\hline
Dimension & Reusable judgments and operations & Intended change in research behavior \\
\hline
Market mechanism & Form competing explanations from participant behavior, constraints, and information diffusion & Move from raw formula variation toward competing market explanations that can be distinguished by evidence \\
Measurement and representation & Check what fields, reference frames, time scales, and transformations actually measure & Correct proxies, scales, and representations so that expression changes are not mistaken for new mechanisms \\
Current portfolio gaps & Use members, weights, and feedback to identify missing information in the current portfolio & Select research questions that may add information not yet covered by the portfolio \\
Test design & Design controls and counterexamples that distinguish competing explanations & Use limited experiments to obtain discriminative evidence so that both successes and failures update judgments \\
Belief updating & Revise the scope of a judgment using supporting evidence and counterexamples & Retain, narrow, revise, disable, or re-enable existing judgments \\
Attention and budget & Allocate effort among exploitation, exploration, measurement correction, and historical review & Avoid prolonged spending on low-information branches while preserving useful depth \\
Engineering support & Reusable retrieval, alignment, validity checking, bounded comparison procedures, and tools & Reduce repeated work and execution errors while supporting research judgments \\
\hline
\end{tabular}

\end{center}
\end{table*}

The Evolving agent carries reusable research resources into later work through skills, procedural documents, and tools. The current-strategy skill stores indexes of the resources currently in use. Other skills organize the questions, evidence, and failure conditions needed for research judgments. Procedural documents store operational steps that can be reused across experiments. Tools implement checkable retrieval, alignment, diagnostics, or bounded comparisons. These saved research artifacts may point to detailed evidence in the research history. Their value is evaluated through later use and resulting portfolio outcomes.

For checkpoint replacement, $ \mathrm{Cap} $ consists of the \texttt{.agents/skills}, \texttt{procedure}, and \texttt{tools} directories in the workspace, together with capability-version records saved by the runtime. Conversation history and historical working files such as \texttt{memory}, \texttt{research}, and \texttt{scratch} belong to $ \mathrm{Hist} $ and are identical at the starting point of both branches. After branching, each branch updates its own history and portfolio. This comparison estimates the subsequent value of accumulated $ \mathrm{Cap} $ relative to the initial $ \mathrm{Cap} $.

Capability updates follow a sequence of research event, resource formation, timely retrieval, action alignment, and evidence-based revision. A real research event triggers the smallest necessary update. A resource retrieved before a relevant decision must state what action it is expected to change. Later actions are then aligned with platform feedback, resource use, and portfolio changes. New evidence determines whether the corresponding content is retained, narrowed, revised, disabled, or re-enabled. Loading a version only means that the resource entered the context. It counts as behavioral evidence only when the later decision changes in the intended way.

\subsection{Data}

This section defines the market-data source, point-in-time universe, 10-minute samples, and temporal boundaries used for evaluation. All processing follows one rule: asset selection, features, and labels at time \(t\) use only information visible within the corresponding evaluation boundary.

\paragraph{Data Source and Dynamic Universe.}
We use the public Binance Data Vision archive of spot-market klines\footnote{\url{https://data.binance.vision/}} as the only raw market-data source. The universe contains USDT-denominated spot trading pairs, with all timestamps in UTC. Let \(T_m\) be the first day of month \(m\), and let \(A_{i,\tau}\) be the USDT turnover of trading pair \(i\) at minute \(\tau\). We measure liquidity using turnover over the previous 30 days and define the monthly universe \(U_m\) as
\begin{equation}
\operatorname{Turnover}_{i,m}=\sum_{\tau\in[T_m-30\mathrm{d},T_m)} A_{i,\tau},
\qquad
U_m=\operatorname{Top60}_{i}\,\operatorname{Turnover}_{i,m}.
\end{equation}
The ranking window ends at \(T_m\) and uses only turnover observed before the new month begins. A candidate trading pair must have at least 30 days of history, appear on at least 20 calendar days in the lookback window, and have at least 95\% one-minute coverage. Stablecoins, leveraged tokens, and other special assets are excluded by rules fixed in advance. \(U_m\) remains unchanged within month \(m\), producing a point-in-time universe that does not use future turnover.

\paragraph{Ten-Minute Bars and Prediction Targets.}
A 10-minute bar is the evaluator's basic time unit and is aggregated from ten one-minute bars in the same UTC window. Open is the first value, high and low are extrema, close is the last value, volume and turnover are summed, and volume-weighted average price is turnover divided by volume. If any source minute is missing, the corresponding asset--bar observation is marked missing. Let \(C_{i,t}\) be the 10-minute close of asset \(i\) at time \(t\). The prediction target is the next-bar close-to-close return:
\begin{equation}
y_{i,t}=\frac{C_{i,t+1}}{C_{i,t}}-1,\qquad i\in U_t.
\end{equation}

\paragraph{Temporal Split and Data Access.}
The development period runs from 2025-01-01 through 2025-12-31, and the OOS test period runs from 2026-01-01 through 2026-06-30. OOS data and results are hidden from the research agent. OOS scoring keeps each portfolio's development-period members, directions, and weights fixed and uses OOS labels only to compute its reported IC. Each period is read from a data snapshot ending on its own final date, and labels near the endpoint also respect the corresponding data boundary. Qlib provides the underlying trading calendar, dynamic instrument lists, and binary market-data access\citep{yang2020qlib}. The AlphaGen design is used for upper-level factor-tensor execution and fixed-capacity factor-pool evaluation.

The following statistics define the nominal number of asset--bar observations as \(60\) times the number of calendar bars and measure missing market-data positions relative to that total. A market observation is considered valid when all six fields---open, high, low, close (OHLC), volume, and turnover---are finite.

\begin{table*}[t]
\caption{Binance Spot dynamic-U60 dataset statistics. Valid assets per bar are reported as minimum/median/maximum.}
\label{tab:crypto-u60-data}
\centering
\small
\begin{tabular}{lrrrr}
\toprule
\multicolumn{5}{l}{\textbf{A. Sample Size and Dynamic Universe}} \\
\addlinespace[2pt]
Period & 10-min bars & Valid assets/bar & Unique assets & Nominal asset--bars \\
\midrule
Development & 52,560 & 58/60/60 & 163 & 3,153,600 \\
OOS test & 26,064 & 59/60/60 & 121 & 1,563,840 \\
Total & 78,624 & 58/60/60 & 199 & 4,717,440 \\
\bottomrule
\end{tabular}
\par\vspace{0.65em}
\begin{tabular}{lrrr}
\toprule
\multicolumn{4}{l}{\textbf{B. Data Completeness}} \\
\addlinespace[2pt]
Period & Missing market positions & Market coverage & Label coverage \\
\midrule
Development & 9,864 & 99.6872\% & 99.6852\% \\
OOS test & 540 & 99.9655\% & 99.9616\% \\
Total & 10,404 & 99.7795\% & 99.7768\% \\
\bottomrule
\end{tabular}

\end{table*}

\subsection{Factor Expressions and Factor-Pool Evaluation}

This section describes how candidate factors are represented, how the factor pool evaluates them, and how fixed-capacity updates create pressure for continued improvement. The expression system, portfolio evaluation, and update rule follow AlphaGen\citep{yu2023generating}. EverMine generates candidate factors and uses platform feedback to organize subsequent search.

\paragraph{Factor Expressions.}
A factor expression is a function from current and historical market fields to a current cross-sectional signal. Its input and output types are defined by the expression tree:
\begin{equation}
\bm{x}^{(k)}_t=f_{\theta_k}\!\left(\{O,H,L,C,V,A,\mathrm{VWAP}\}_{\le t}\right)
\in\mathbb{R}^{|U_t|}.
\end{equation}
Leaf nodes are market fields or constants. Internal nodes include elementwise unary and binary operations, rolling time-series operators, two-series rolling operators, conditional operations, and cross-sectional ranking at the same timestamp. Nodes can be nested subject to type rules. Rolling windows range from 1 to 1024 bars. \(\operatorname{Ref}(x,n)\) with \(n>0\) reads only past values, and \(\operatorname{CSRank}\) is computed within the current \(U_t\).

\paragraph{Cross-Sectional Normalization and Ensemble Scores.}
At each timestamp, the evaluator separately standardizes factor \(k\) and the label within \(U_t\), producing \(\widetilde{\bm{x}}^{(k)}_t\) and \(\widetilde{\bm{y}}_t\). Means and population standard deviations use only finite values, and missing positions are filled with zero after standardization. Given \(K\) current members and signed weights \(\bm{w}\), the portfolio signal and portfolio Information Coefficient (IC) are
\begin{equation}
\bm{s}_t(\bm{w})=\sum_{k=1}^{K}w_k\widetilde{\bm{x}}^{(k)}_t,
\qquad
\operatorname{IC}(\bm{w})=\frac{1}{|\mathcal{T}|}\sum_{t\in\mathcal{T}}
\operatorname{Corr}_{i\in U_t}\!\left(s_{i,t},\widetilde y_{i,t}\right).
\end{equation}
Thus, \(\operatorname{IC}(\bm{w})\) is the time average of per-bar cross-sectional Pearson correlations and is the main quality measure for the factor pool. Single-factor IC, Rank IC, and coverage are stored as diagnostics. Candidate admission is determined by the joint factor-pool update.

\paragraph{Joint Fitting and Fixed-Capacity Updates.}
The joint update tests whether a new candidate adds information to the current portfolio. Let \(c_k\) be the mean cross-sectional Pearson correlation between factor \(k\) and the label, and let \(M_{jk}\) be the mean cross-sectional Pearson correlation between factors \(j\) and \(k\). After adding a candidate, the AlphaGen factor pool refits all members by
\begin{equation}
\bm{w}^{*}=\arg\min_{\bm{w}}
\left(\bm{w}^{\mathsf T}M\bm{w}-2\bm{c}^{\mathsf T}\bm{w}+1\right)
+\lambda\lVert\bm{w}\rVert_1,
\qquad \lambda=0.005.
\end{equation}
A candidate is rejected directly if its signed mean Pearson correlation with any existing member is strictly greater than 0.99. We impose no minimum single-factor IC. The factor-pool capacity is fixed at 30. If the refitted pool exceeds this capacity, the member with the smallest \(|w_k|\) is removed. If the candidate itself is removed, the update fails; otherwise, the update produces a member replacement. Weights may be negative. The development-task target of \(0.25\) defines only a success endpoint for long-horizon research and does not affect candidate admission.

\subsection{Compute Environment and Common Conditions}

Each agent sandbox has 6 CPUs, 15 GiB of memory, and a 40 GiB workspace. Candidate-factor quality is evaluated on a machine with four NVIDIA RTX 4090 GPUs, with one GPU assigned per request. Evaluation uses deterministic CUDA settings and float32 factor-tensor execution. Local LLM inference runs on a machine with eight NVIDIA A800 80GB GPUs. The full experimental suite can be completed within two weeks on this hardware.

\subsection{Multi-Resource Budgets and Continuous Research Progress}

Resource budgets are listed in Section~\ref{sec:experimental-setup}. Complete long trajectories additionally allow up to 10 recoveries from agent-resource failures and 3 recoveries from transient platform failures. Continuation branches allow 2 and 3 such recoveries, respectively. The terminal result is the last fully evaluated, deliverable factor portfolio available when research stops. We also record the stopping criterion, all resource use, and the termination reason.

\subsubsection{Joint-Budget Progress and Checkpoints}

The two capability-replacement conditions start with the same resource-consumption record. Only the additional continuation budgets listed in the main text are metered after branching.

The multi-resource budget constrains several forms of research effort that are not interchangeable. To compare how far a trajectory has progressed, we select six continuously accumulated resources,
\[
\begin{aligned}
\mathcal{J}=\{&\text{model cost, sandbox CPU, platform submissions},\\
&\text{output tokens, successful research responses, cumulative writes}\}.
\end{aligned}
\]
and define continuous research progress for run \(i\) at time \(t\) as
\begin{equation}
B_i(t)=\max_{j\in\mathcal{J}}\frac{C_{ij}(t)}{L_j},
\end{equation}
where \(C_{ij}(t)\) is cumulative use of resource \(j\) and \(L_j\) is its fixed limit. \(B_i(t)\leq b\) means that all six cumulative resources remain within the same fraction \(b\) of their full limits. The runtime saves analysis checkpoints when \(B\) first reaches 25\%, 50\%, and 75\%. The 25\% and 75\% checkpoints also save the parent-trajectory states needed for capability replacement. Saving a checkpoint does not interrupt the agent or change its later actions.

\(B\) represents progress through the joint budget only; it does not assign exchange rates between resources. Process results are also reported on actual resource axes such as model cost, CPU core-hours, submissions, output tokens, successful research responses, and active wall-clock time. Two checkpoints can have the same \(B\) while having different resource compositions, and those compositions are retained. Runs are compared on the same \(B\) axis only when they share the same resource set and fixed limits. Different budgets and different models are reported separately.

\subsubsection{Model-Cost Accounting}

We meter model usage with a frozen reference-cost ledger. For uncached input, cache-read input, cache-write input, and total generated output token counts $U,H,W,O$, respectively, the equivalent model cost is
\begin{equation}
C_{\mathrm{USD}}=\frac{U p_u+H p_h+W p_w+O p_o}{10^6},
\end{equation}
where the prices are in USD per million tokens and $O$ includes reasoning tokens. 

We use representative official commercial prices when they are available: the DeepSeek card is its official API price, $(p_u,p_h,p_w,p_o)=(0.15,0.003,0.15,0.60)$, with cache usage reported by the provider.

Qwen3.8-27B belongs to Qwen's open-weight series, while Alibaba Cloud's commercial serving and optimization focus on the Flash, Plus, and Max tiers. Its hosted-API list price is therefore relatively high for the local deployment pattern commonly used for this model in academic and industrial research. We estimate a deployment-equivalent Qwen price card from the historical utilization cost of a four-A800 low-priority allocation. At RMB~1.60 per A800-hour, four cards over the 24.118896-hour calibration window cost RMB~154.360934. Pricing the same audited workload at the official Qwen3.8-27B rates gives RMB~266.331898, yielding a ratio of 0.579581; we round it to 0.6 and apply it to the official token rates. The resulting frozen card is $(p_u,p_h,p_w,p_o)=(0.265710,0.053142,0.265710,1.062840)$, using a 95\% cache-hit split and an exchange rate of RMB~6.7743 per USD. These input and output rates are close to the official Qwen3.8-Plus API tier, providing an external price-scale check.

The ledger accumulates every usage record emitted by the model service, including retry attempts and failed requests that emit usage; absent usage raises a measurement failure. Each model's price card and accounting assumptions are fixed across conditions and inherited unchanged by continuation branches.

\subsection{Trajectory Alignment under Heterogeneous Stopping Constraints}

When a run stops because of model cost, CPU, submissions, output tokens, successful research responses, or cumulative writes, the limiting resource is already included in \(\mathcal{J}\), so terminal \(B\) is usually close to 100\%. Active wall-clock time is affected by model-service and evaluator queues. We therefore treat it separately as a measure of delivery speed and do not let it advance \(B\). A run that reaches its wall-clock limit retains its actual \(B\) and is marked as censored by the time boundary. Memory and workspace usage are capacities, while recovery counts are infrequent safety events, so they are also excluded from continuous research progress.

Process comparisons use only the range actually covered by each run. At fixed \(B\) checkpoints, we compare portfolio quality \(Q(B)\). Resource efficiency and delivery speed are then compared at matched positions on each resource axis. Portfolio quality updates as a step function only after an evaluator response is completed. Between evaluations, we carry forward the most recent deliverable portfolio quality. After the final evaluation, the last horizontal segment extends only to the true endpoint of that run. Missing checkpoints or resource positions are left missing: we do not interpolate, extrapolate, or stretch terminal states to 100\%. Checkpoint estimates use only runs in each condition that actually reached the checkpoint and report the number of covered trajectories.

Stopping at a planned resource limit, producing no new resource, making no valid submission, and leaving resources unused are all retained as system outcomes. Infrastructure failures are marked separately as platform censoring. Replacement runs are launched only under rules fixed before inspecting results, and the original trajectories and their realized costs remain in the reliability report.

\section{Additional Analyses and Results}
\label{app:analysis-details}

\subsection{Endpoint and OOS Reporting}

The development-period and OOS effects in the end-to-end evaluation use the final deliverable factor pool as the unit of outcome. Fixed and Evolving are independent groups of research runs within each model. Table~\ref{tab:primary-effects} reports the number of runs in each group, the arithmetic mean of portfolio Pearson IC, the mean difference of Evolving minus Fixed, and its interval. Table~\ref{tab:capability-set-replacement-effects} reports subsequent gains under the two $ \mathrm{Cap} $ states from the same starting state and their difference. OOS endpoint evaluation measures the final research product on the held-out period, while development-period trajectories are used to analyze the research process.

OOS endpoints are available for all 18 end-to-end trajectories and all 24 paired blocks in the capability-state replacement experiment. End-to-end intervals resample independent trajectories. For capability replacement, we first average the two repetitions within each parent trajectory, then weight parent trajectories equally and resample them. Matching run IDs are bookkeeping labels; the two end-to-end groups are compared as independent samples.

\begin{figure}[t]
\centering
\includegraphics[width=\linewidth]{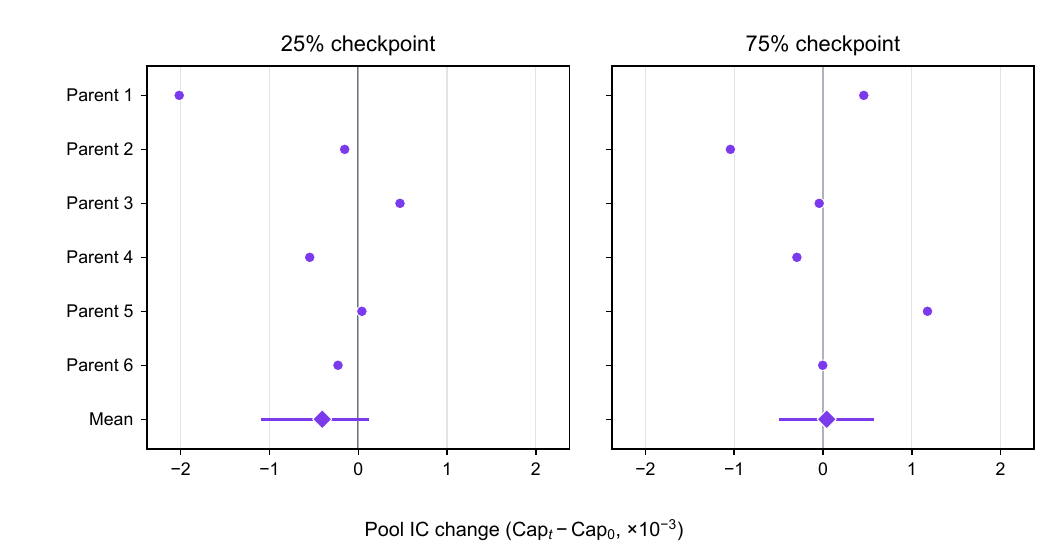}
\caption{\textbf{Capability-state replacement effects by parent trajectory.} Dots show parent-trajectory effects and diamonds show means. Positive values indicate larger subsequent improvement under \(\mathrm{Cap}_t\).}
\label{fig:capability-set-replacement-effects}
\end{figure}

Development-period effects by parent trajectory show two positive and four negative parent effects at each checkpoint. Across the 12 parent-trajectory--checkpoint pairs, the two repetitions have opposite signs in five cases. The cost of forming $ \mathrm{Cap} $ has already been incurred before the checkpoint, while continuation branches still incur the costs of reading, executing, and recovering capability resources.

\subsection{Trajectory and Resource Robustness}

Figure~\ref{fig:long-horizon-performance} in the main text divides each trajectory's cumulative submission count \(S_i(t)\) by its final submission count \(S_i(T_i)\) and maps the result to 0--1000. This axis aligns relative positions in the submission sequence. Absolute submission counts and resource use are reported separately.

\begin{table}[h]
\caption{Development-period factor-pool IC at selected joint-budget checkpoints. Parentheses show the number of trajectories reaching each checkpoint.}
\label{tab:joint-budget-quality}
\begin{center}
\small
\resizebox{\linewidth}{!}{
\begin{tabular}{llccc}
\toprule
Model & Condition & 20\% & 50\% & 80\% \\
\midrule
Qwen3.8-27B & Fixed & 0.06013 (6) & 0.06066 (6) & 0.06090 (5) \\
Qwen3.8-27B & Evolving & 0.05898 (6) & 0.06039 (6) & 0.06149 (6) \\
\addlinespace
DeepSeek-V4.1-Flash & Fixed & 0.06319 (3) & 0.06431 (3) & 0.06618 (3) \\
DeepSeek-V4.1-Flash & Evolving & 0.06216 (3) & 0.06382 (3) & 0.06442 (3) \\
\bottomrule
\end{tabular}
}
\end{center}
\end{table}

Table~\ref{tab:joint-budget-quality} adds three process checkpoints from the joint-budget view. The 20\% and 50\% checkpoints include all 18 trajectories. At 80\%, only the Qwen3.8-27B Fixed group is missing one trajectory. For Qwen3.8-27B, the Evolving mean is slightly higher than the Fixed mean at 80\%. For DeepSeek-V4.1-Flash, the Fixed mean is slightly higher at all three checkpoints. The table reports group means at joint-budget checkpoints, while the main figure reports group medians over relative submission progress. These views show resource progress and within-trajectory progress, respectively.

\begin{figure*}[t]
    \centering
    \includegraphics[width=\textwidth]{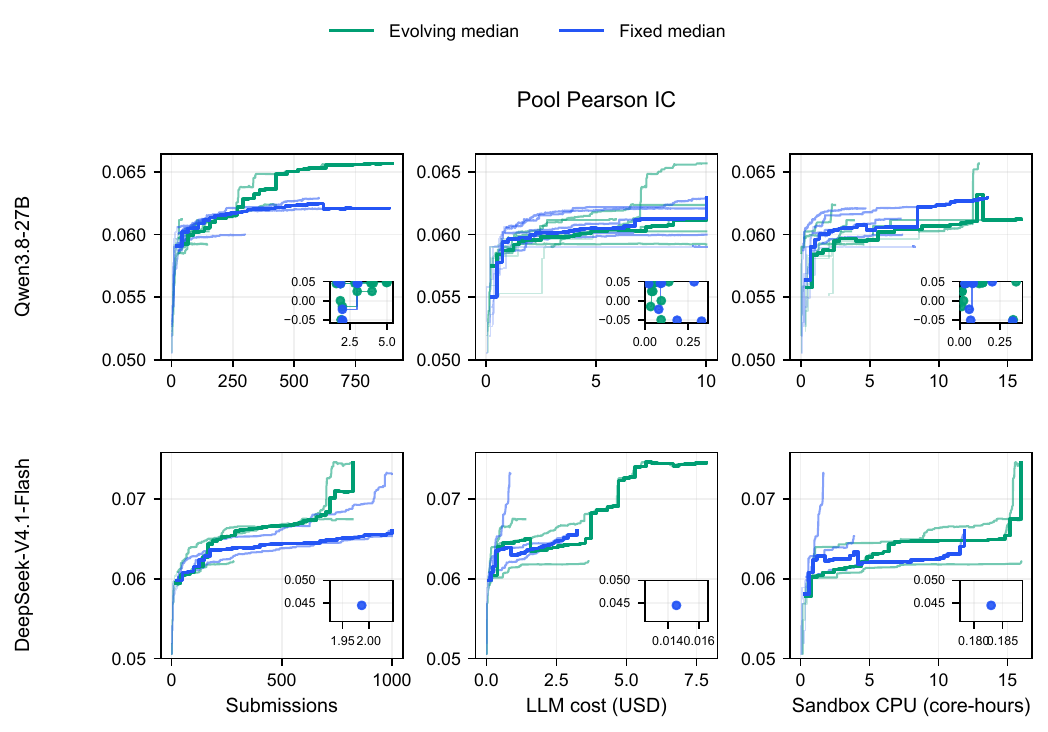}
    \caption{\textbf{Long-horizon research trajectories aligned by matched resource consumption.} Thin lines are individual trajectories, thick lines are group medians, and crosses mark censored endpoints.}
    \label{fig:resource-sensitivity-curves}
\end{figure*}

Comparisons aligned by matched resource use preserve two main patterns from the primary trajectory figure. Absolute portfolio quality forms quickly early in research, and later differences are driven mainly by a small number of long-tail trajectories. Figure~\ref{fig:resource-sensitivity-curves} also shows that the observable length of a trajectory depends on the resource axis. We therefore use these views as process diagnostics. Final portfolio quality under the common resource limits remains the endpoint result reported in the main text.

\subsection{Capability-State Behavioral Diagnostics}
\label{app:cap-behavior}

\paragraph{$ \mathrm{Cap} $ content and rule revision.}
We inspect $ \mathrm{Cap} $ files for 12 parent-trajectory checkpoints and all 48 continuation branches. Branches that start from accumulated $ \mathrm{Cap} $ can still revise their research judgments in response to new feedback through $ \mathrm{Hist} $, while the loaded $ \mathrm{Cap} $ snapshot remains read-only. In one Qwen Evolving parent trajectory (record e11), a continuation from the 75\% checkpoint submits a volume-weighted-average-price recovery indicator that had previously been expected to have low marginal value. The factor produces a gain of \(+0.00003067\). The branch then narrows the scope of the earlier judgment and records the revision in its research directory, while the loaded $ \mathrm{Cap} $ snapshot itself stays unchanged.

\paragraph{Concrete choices from a shared state.}
Another Qwen Evolving parent trajectory (record e16) produces four continuation branches from the 75\% checkpoint. Before the first new submission, all four branches have identical portfolio members, weights, and versions. Two candidate factors represent short-term price declines and prices below a recent volume-weighted close. The first takes the minimum one-bar close return over the past 72 bars and then applies a cross-sectional rank. The second checks whether both the current low and close are below the volume-weighted close over the past 12 bars, and then ranks the resulting indicator cross-sectionally. One branch with initial $ \mathrm{Cap} $ selects the first candidate and obtains \(-0.00002382\). The other branch with initial $ \mathrm{Cap} $ and both branches with accumulated $ \mathrm{Cap} $ select the second candidate and each obtain \(+0.00001915\). Their local screening ICs are \(0.00891\) and \(0.01768\), with corresponding member correlations of about \(0.814\) and \(0.575\). This checkpoint links visible evidence, realized choices, and portfolio outcomes while also showing choice variation within the same $ \mathrm{Cap} $ condition.

\paragraph{Late-stage persistence.}
An analysis of portfolio outputs that does not use return labels covers 40 branches: all six parent trajectories at the 25\% checkpoint and five parent trajectories at the 75\% checkpoint. The correlation between each final output and its own starting portfolio is \(0.9700 / 0.9755\) under initial / accumulated $ \mathrm{Cap} $ at the early checkpoint and \(0.9951 / 0.9924\) at the late checkpoint. Stronger late-stage persistence therefore appears under both $ \mathrm{Cap} $ conditions. The stage comparison reflects the joint evolution of research history, the portfolio, and capability state.

\begin{figure*}[t]
    \centering
    \includegraphics[width=\textwidth]{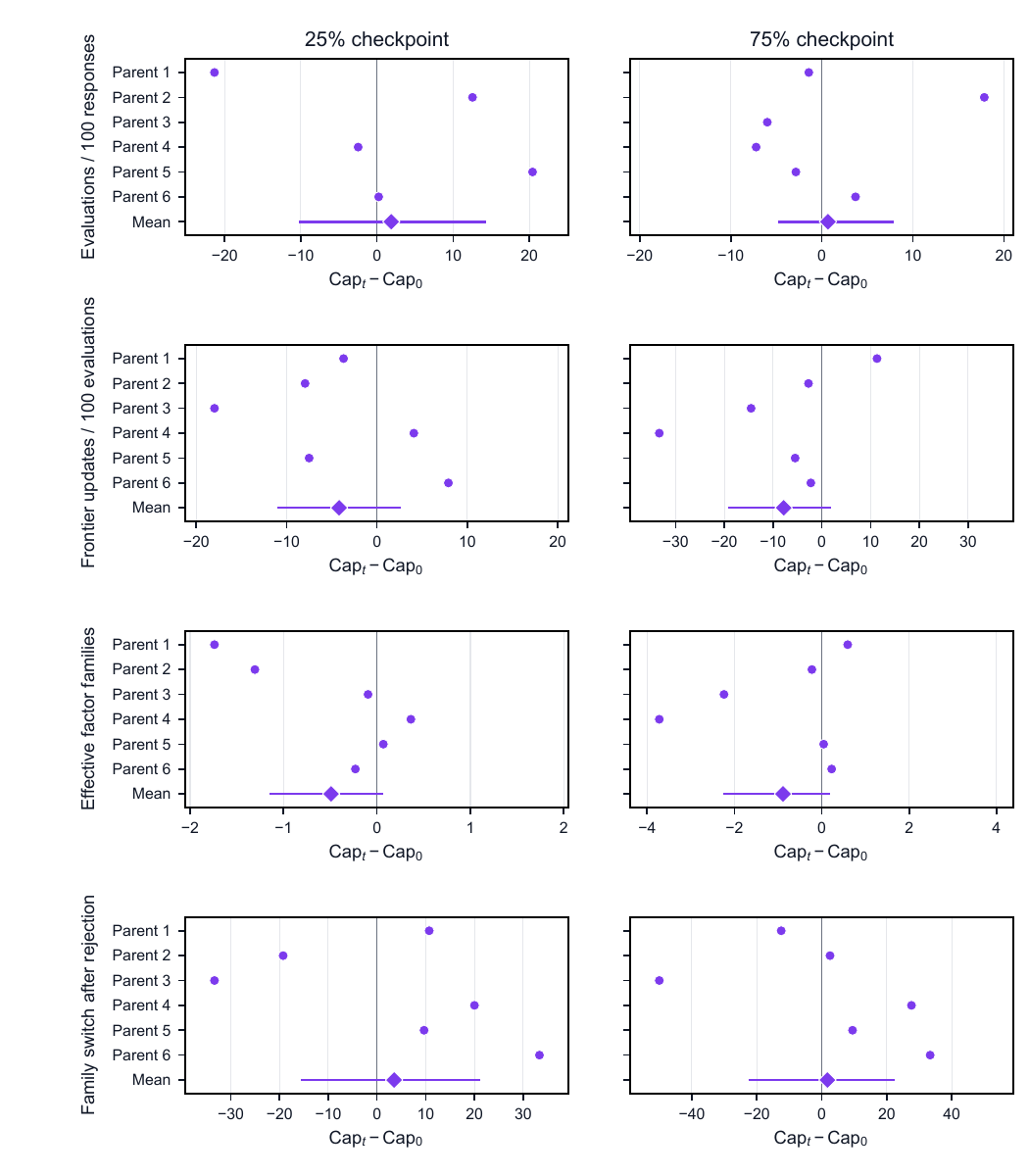}
    \caption{\textbf{Behavioral effects of capability-state replacement.} Dots show means across repetitions within each parent trajectory. Diamonds and horizontal lines show means across parent trajectories and 95\% bootstrap intervals.}
    \label{fig:e5-behavior-effects}
\end{figure*}

Factor families are classified from the data fields and operators used in each expression. The effective number of factor families is the exponential of the entropy of the submission-frequency distribution across families. It reflects both the number of families and the balance of submissions across them. The post-rejection family-switch rate is the fraction of rejected submissions for which the next submission belongs to a different factor family, among rejection events that have a subsequent submission.

Figure~\ref{fig:e5-behavior-effects} reports four behavioral metrics in their original units. At the 25\% checkpoint, the mean \(\mathrm{Cap}_t-\mathrm{Cap}_0\) differences are \(-0.490\) for the effective number of factor families, \(+1.878\) evaluations per 100 model responses, \(-4.192\) percentage points for the share of submissions that improve portfolio IC, and \(+3.523\) percentage points for the post-rejection family-switch rate. At the 75\% checkpoint, the corresponding differences are \(-0.888\), \(+0.679\), \(-7.829\), and \(+1.734\). Depending on metric availability, each row covers 5--6 parent trajectories and 9--12 paired blocks. All intervals include zero.

\subsection{Research Behavior Evidence}
\label{app:decision-evidence}

We use submission records, original trajectory text, and replay results to verify the portfolio gains associated with parameter tuning, screening, and reevaluation. This section describes the covered trajectories, the basis for candidate-selection decisions, and the procedure used to restore historical platform states.

\subsubsection{Submission Categories and Contributions}

Using prior submission records, we distinguish new structures, parameter tuning of existing structures, and exact resubmissions. The analysis covers 8,215 submissions that occur strictly after the factor pool first reaches capacity across the 18 trajectories. Marginal gains are available for 8,211 of them. Classification uses only history available before each submission. We represent an expression tree by retaining fields, operators, and their order while abstracting numerical constants. This representation is used to compare structures. Exact-formula comparison also retains numerical values while ignoring whitespace and superficial parenthesis formatting. A new-structure submission uses a structure that has not appeared previously in the same trajectory. A parameter-tuning submission keeps an existing structure but changes numerical parameters, producing a formula that has not previously been submitted. An exact resubmission uses a formula that has already been submitted, even if the factor ID differs. ``Previously recorded'' always refers to prior platform submissions in the same trajectory. This classification describes expression construction; economic mechanism and signal direction are evaluated separately.

After the pool first reaches capacity, there are 5,075 new-structure submissions, 2,318 parameter-tuning submissions, and 822 exact resubmissions. The four missing outcomes all belong to the resubmission category. We report known contributions and keep these observations marked as missing. For each trajectory, we sum realized IC changes within each category and then average equally within groups, retaining all negative values. Positive net contributions from parameter tuning occur in 4/6 Qwen Fixed, 6/6 Qwen Evolving, and 3/3 trajectories in each DeepSeek group. The Qwen Fixed group also contains one negative trajectory and one trajectory with no parameter-tuning attempt. In every group, the mean parameter-tuning contribution remains positive after deleting any one trajectory. Using \(10^{-4}\) as a gain threshold, parameter tuning contributes 44 of the 201 total hits. None of the 818 known exact-resubmission outcomes exceeds this threshold.

\subsubsection{Screening Decisions and Replay}

The replay study began as an exploratory case audit. We verify the screening basis, all valid candidate expressions already written at that point, and each accept-or-skip decision from the original text and scripts before restoring the complete candidate batch from the historical state. The candidate sets include every accepted and screened-out item in the selected batches. Batches A and B in the main text come from the same Qwen Evolving trajectory. Batch A covers the four candidates that completed residual screening in Experiment 014. Batch B covers all eight candidates in the initial screen of Experiment 024. Of these 12 candidates, 11 were initially screened out and one was accepted directly. One screened-out candidate was submitted later with a delay. Counts use the initial decision in the original batch, and gains refer to the state at that decision. Batch selection was post hoc, so all proportions are scoped to these two batches.

The plan for Batch A explicitly reuses the submission rule from Experiment 013. It requires an absolute residual IC of at least \(0.0015\) and also checks correlation with existing members. The residual is obtained by regressing the candidate factor on current pool members and measures predictive correlation after removing the component explained by those members. In the actual decision, the agent compares two representations of the same construction. The standard representation has residual IC \(0.00146\) and falls below the threshold, while the rank representation has residual IC \(0.00395\) and passes. At the time of the decision, registered Procedure Version 2 emphasizes residuals and correlations. Version 3 later makes the \(0.0015\) threshold explicit and uses the rejected standard representation as an execution example. This case records how a rule moves from a specific experiment into \(\mathrm{Cap}\).

The plan for Batch B explicitly cites four checks in Procedure Version 4 from \(\mathrm{Cap}\) and adds a preliminary single-factor IC screen. The plan requires absolute IC of at least \(0.012\) and the same sign in all four quarters. None of the eight candidates meets this numerical threshold, so the agent stops before residual screening and submission. Registered Procedure Version 4 retains residual and correlation checks, while the specific \(0.012\) preliminary threshold comes from the current experiment plan. We verify rule provenance jointly from version contents, the experiment plan, and the actual decision, using registration times to establish order.

For single-candidate replay, we restore the original state separately for each factor. We preserve member order, weights, evaluator internal state, and failure-record caches, and use the same evaluator version and development data. Two historically evaluated candidates are each replayed twice; all four control evaluations match the original records. The ten previously unevaluated candidates are each run once. Among screened-out candidates, four have positive gains, six have negative gains, and one has a zero gain. All four positive values exceed \(10^{-5}\), and one exceeds \(10^{-4}\). The accepted candidate in Batch A is the best candidate in that batch. These replays measure the candidate's marginal effect on development-period portfolio IC at the original decision state.

A separate Qwen Fixed trajectory provides a complete rescreening case with eight historical formulas. Under the final decision coding, two are screened out and six are accepted; one candidate moves from deferred to accepted. Each candidate is evaluated twice from the exact state of its decision, and all control and repeated evaluations agree. All eight candidates are admitted through member replacement, but all gains are negative. The two screened-out factors have gains of \(-0.00000034645\) and \(-0.00002667680\). Across the three batches from two trajectories, there are 20 candidates and 13 screened-out candidates. All four screened-out candidates with positive gains at the original portfolio come from the same Evolving trajectory. This coverage therefore includes both missed-opportunity and no-missed-opportunity cases.

Sequential replay preserves the full platform update after every submission. The candidate sets and submission order are frozen before sequential evaluation. In Batch A, both paths first submit the originally accepted candidate. The additional-submission path then submits the three screened-out candidates. In Batch B, we compare the unchanged original portfolio against a path that evaluates all eight candidates in their original order. The complete output state after each step becomes the input to the next step, preserving rejections, negative gains, and cache updates. Control evaluations from the historical records and two repetitions of each sequential step agree. The final endpoint differences are \(-0.00001749396\) and \(-0.00000666454\), with three and eight additional evaluations, respectively. Among the 11 screened-out candidates, seven strictly change the sign of their marginal gain between independent evaluation on the original portfolio and their position in the sequence, with magnitudes above \(10^{-6}\) on both sides. Three of the four candidates that are positive on the original portfolio become negative in sequence. Capacity constraints, weight refitting, and evaluator state all participate in these updates. The replay therefore reports the final result under a submission order fixed in advance.

\subsubsection{Reevaluation and Rule Revision}

We compare the marginal gain of the same expression across two consecutive submissions. Within each trajectory, we match consecutive submissions of the same expression and retain pairs for which the pool is already full at both submissions. This produces 781 pairs across 15 trajectories, with usable gains for both submissions in 777 pairs. Let \(\epsilon=10^{-6}\). A gain below \(-\epsilon\) is negative and a gain above \(\epsilon\) is positive. Among 178 pairs whose first gain is negative, 50 become positive on resubmission. Among 311 pairs whose first gain is positive, 79 become negative. Of the 50 negative-to-positive cases, 44 come from Qwen Fixed. The marginal value of an old factor can therefore move in both directions as the portfolio changes. This observation covers factors that agents actually chose to resubmit.

We also examine how agents revise existing rules after new feedback. In one DeepSeek Evolving trajectory, the agent generalizes a previous failure into a local residual-screening threshold of about \(0.01\) and stops submitting one-period-lagged reversal candidates. The agent later queries the shared research-status interface. Its automatically generated response restates the pre-specified rule that valid hypotheses can be submitted without first passing a local metric; this response is triggered by the status query, not by the trajectory's screening behavior (Appendix~\ref{app:research-protocol}). The agent then submits the candidate on the unchanged pool version. Portfolio IC rises from \(0.06015955\) to \(0.06026769\), after which the agent withdraws the hard threshold. This is a delayed evaluation and rule revision from the same portfolio state, and the gain contributes to the realized endpoint. The case combines a protocol-compliance reminder with counterevidence from the submitted candidate, so it does not isolate autonomous revision in response to evaluation feedback alone. We verify the case from the screening rationale in the original trajectory, the status response, the actual submission, and the revision after feedback.

\subsection{Factor-Pool Similarity and Diversity}
\label{app:factor-pool-analysis}

The factor-pool analysis tests whether differences in research behavior appear in the final retained signals. We run every final factor pool on the same set of market segments fixed in advance from 2025 and standardize each factor output cross-sectionally over the dynamic universe. These market segments do not use return labels. Portfolio quality is read from the standard end-to-end results and reported alongside geometric measures.

\subsubsection{Geometric Measures and Comparison}

For a factor pool with \(K\) valid signal directions, let \(R\) be the correlation matrix of factor outputs on the fixed market segments and \(\lambda_k\) its eigenvalues. Define normalized spectral weights as \(p_k=\lambda_k/\sum_j\lambda_j\). The effective signal dimension is
\[
D_{\mathrm{eff}}=\exp\!\left(-\sum_{k=1}^{K}p_k\log p_k\right).
\]
We also report mean absolute within-pool correlation, the number of near-duplicate directions, and the number of degenerate factors. Between two pools, we use a one-to-one optimal matching distance based on absolute correlation. Correlation between portfolio outputs measures similarity between the final combined signals.

The equal-reference nearest-neighbor distance finds, for each factor, the closest signal in one Fixed reference pool. The distance is \(\sqrt{1-\rho^2}\), where \(\rho\) is the correlation between the two signals. We first average equally across factors in the query pool and then across same-model Fixed reference pools. When a Fixed pool is the query, it is excluded from its own reference set. Both conditions are compared against one reference pool at a time. A larger distance means that the factor lies farther from any individual signal direction in the reference pool.

The subspace residual linearly reconstructs each query factor from the signals in one Fixed reference pool and measures the unexplained squared residual of the standardized signal. We then average equally across factors and reference pools. Leave-one-quarter-out reconstruction error fits reconstruction coefficients on the other quarters and computes mean squared reconstruction error on the held-out quarter, then averages across quarters, factors, and reference pools. The first measure captures how much of a query signal is not explained by linear combinations of the reference signals. The second tests whether the reconstruction relationship remains stable across quarters. Larger values indicate more unexplained signal components and larger cross-quarter reconstruction errors, respectively.

We compare Fixed and Evolving separately within each model. The independent statistical unit is the complete factor pool from one research trajectory. Condition differences are tested by exact permutation of complete factor-pool labels. Individual members and two-dimensional projections are used descriptively to show within-pool composition. A final pool reflects both candidate generation and fixed-capacity selection, so these geometric results are interpreted together with submission behavior and portfolio quality.

\subsubsection{Overall and Family-Level Results}

Across the 18 final factor pools, changes in signal dimension and between-pool distance differ by model. For Qwen3.8-27B, Evolving minus Fixed is \(+0.630\) for effective dimension (exact permutation \(p=0.649\)), \(+0.0413\) for equal-reference nearest-neighbor distance (\(p=0.214\)), \(+0.0520\) for subspace residual (\(p=0.292\)), and \(+0.0526\) for leave-one-quarter-out reconstruction error (\(p=0.290\)). For DeepSeek-V4.1-Flash, the corresponding differences are \(-2.147\) (\(p=0.500\)), \(-0.0322\) (\(p=0.450\)), \(-0.0672\) (\(p=0.450\)), and \(-0.0723\) (\(p=0.450\)). The three outward-distance measures increase slightly for Qwen and decrease for DeepSeek. None of the overall within-model differences is statistically significant.

Family affinity first computes the maximum absolute correlation between each factor and a set of reference expressions for one signal family, then averages equally across factors in the pool. A larger value means that pool members are, on average, closer to reference signals in that family. The same factor may be related to more than one family.

Exploratory family coordinates show changes in composition. For Qwen, short-horizon price-change affinity decreases by \(0.1021\) (\(p=0.0043\)), price-volume coupling affinity decreases by \(0.0559\) (\(p=0.0455\)), and volatility-and-range affinity increases by \(0.0278\) (\(p=0.268\)). For DeepSeek, affinity increases slightly for all five families, with the largest increase in volatility and range at \(0.0668\) (\(p=0.100\)). This family dictionary was introduced as an exploratory coordinate system during endpoint analysis to describe the types of signals retained in the final pools. Portfolio predictive performance is reported by the end-to-end results.

\subsection{Realized Resource Use}
\label{app:realized-compute}

Resource use differs across models. DeepSeek Fixed / Evolving use an average of \(5.8021 / 16.0003\) CPU core-hours and USD \(2.3403 / 4.2973\) equivalent model cost, and complete \(1000 / 642\) evaluations. All three Fixed runs stop at the submission limit, while all three Evolving runs stop at the CPU limit. For Qwen, the corresponding averages are \(7.2205 / 7.2329\) CPU core-hours, USD \(9.4538 / 9.3240\) model cost, and \(346.0 / 295.3\) evaluations. Five runs in each condition stop at the model-cost limit; the remaining Fixed and Evolving runs stop at the wall-clock and CPU limits, respectively. The capability-state replacement experiment completes 4,126 evaluations. The mean \(\mathrm{Cap}_t-\mathrm{Cap}_0\) CPU differences are \(-0.13108 / +0.02117\) core-hours at the early / late checkpoints, and the evaluation-count differences are \(+10.17 / +1.33\). Leave-one-parent-trajectory-out means cross zero in all cases. Resource ledgers record cumulative costs for complete trajectories, and we use them to compare total resource consumption between conditions.

\end{document}